\documentclass[lettersize,journal]{IEEEtran}
\usepackage{amsmath,amsfonts}
\usepackage{algorithmic}
\usepackage{algorithm}
\usepackage{array}
\usepackage[caption=false,font=normalsize,labelfont=sf,textfont=sf]{subfig}
\usepackage{textcomp}
\usepackage{stfloats}
\usepackage{url}
\usepackage{verbatim}
\usepackage{graphicx}
\usepackage{cite}
\usepackage{booktabs} 
\usepackage[table]{xcolor}
\usepackage{tabularx}
\usepackage{multirow}
\usepackage{xcolor}
\usepackage[table]{xcolor}
\usepackage{graphicx}
\usepackage{array}
\usepackage{booktabs}
\usepackage{tabularx}
\usepackage{multirow}

\newcolumntype{P}[1]{%
  >{\raggedright\arraybackslash}p{#1}}
\newcolumntype{Y}{%
  >{\raggedright\arraybackslash}X}

\begin{document}

\title{From Recognition to Reasoning: Advancing Multimodal Harmful Meme Detection via Chain-of-Thought Alignment}

\author{
Hexiang~Gu,
Qifan~Yu,
Yuan~Liu,
Zikang~Li,
Saihui~Hou,~\IEEEmembership{Member,~IEEE},
Jian~Zhao,
and~Zhaofeng~He,~\IEEEmembership{Member,~IEEE}%
\thanks{Hexiang Gu and Qifan Yu contributed equally to this work.}
\thanks{Hexiang Gu, Qifan Yu, and Zhaofeng He are with
Beijing University of Posts and Telecommunications, Beijing 100876, China
(E-mail: guhexiang@bupt.edu.cn; 2025111227@bupt.cn;
 zhaofenghe@bupt.edu.cn).}
\thanks{Yuan Liu is with the School of Artificial
Intelligence, Beijing Normal University, Beijing 100875, China
(E-mail: liuyuan2024@mail.bnu.edu.cn).}
\thanks{
Saihui Hou is with the School of Artificial Intelligence, Beijing Normal University, Beijing 100875, China. (Email: housaihui@bnu.edu.cn)
}

\thanks{
Zikang Li is with Beijing University of Posts and Telecommunications, Beijing 100876, China, and also with Zhongguancun Academy, Beijing 100094, China. (Email: zikangli@bupt.edu.cn)
}

\thanks{Jian Zhao is with Zhongguancun Academy,
Beijing 100094, China, and also with Zhongguancun Institute of Artificial Intelligence, Beijing 100094, China.
(E-mail: jianzhao@zgci.ac.cn).}

\thanks{Corresponding authors: Saihui Hou and Zhaofeng He.}
}

\markboth{IEEE Transactions on Information Forensics and Security}%
{Hexiang Gu \MakeLowercase{\textit{et al.}}: MemeMind: A Large scale harmful meme dataset with Chain-of-Thought Annotation }


\maketitle

\begin{abstract}
As a multimodal communication medium that integrates images and text, memes often convey implicit harmful content through metaphors, satire, and humor, making harmful meme detection a complex and challenging task. Although recent studies have achieved considerable progress in detection accuracy and model interpretability, large-scale, high-quality datasets for harmful memes remain scarce. Moreover, existing methods still exhibit notable limitations in identifying implicit risks and understanding fine-grained semantics. To address these challenges, we construct MemeMind, a large-scale dataset for harmful meme detection. MemeMind comprises a broad collection of publicly available memes and adopts a rigorous and comprehensive taxonomy of harmful content developed in accordance with widely recognized international standards and contemporary Internet contexts. In addition, the dataset provides detailed structured Chain-of-Thought (CoT) reasoning annotations to support fine-grained analysis of harmfulness, implicit intentions, and underlying semantics in memes. Building upon MemeMind, we further propose MemeGuard, a reasoning-oriented multimodal framework for harmful meme detection. MemeGuard employs a three-stage training strategy to progressively enhance the model’s visual understanding, multimodal reasoning, and harmful content discrimination capabilities, thereby improving both detection accuracy and the interpretability of model decisions. Extensive experimental results demonstrate that MemeGuard outperforms existing state-of-the-art methods on the MemeMind dataset, providing a solid foundation for future research on harmful meme detection and multimodal content safety.

\end{abstract}


\noindent
{\small\textcolor{red}{\textit{Warning: This paper contains potentially disturbing meme examples involving harmful content, presented solely for research purposes.}}}

\section{Introduction}
With the exponential growth of social media, memes have proliferated as a ubiquitous form of online expression. However, their multimodal nature makes them a powerful vehicle for disseminating harmful content, which may be conveyed either explicitly or implicitly, including hate speech~\cite{ref1,ref2}, discrimination~\cite{ref42}, and violent implications~\cite{ref43}, often in implicit forms. This implicitness, combined with the tight semantic coupling between visual and textual modalities, enables such content to circumvent traditional content moderation systems, making harmful meme detection an increasingly critical challenge in the field of cybersecurity.

To address the above limitation of harmful meme detection, several datasets have been proposed~\cite{ref2,ref3,ref4,ref5,ref6}. However, these datasets still suffer from several major limitations: 1) \textbf{Limited scale.} Existing datasets are generally small, typically containing only a few thousand to around ten thousand samples and covering a narrow range of harmful categories, which restricts the study of model generalization and scalability.
2) \textbf{Inconsistent classification standards.} Due to diverse data sources and variations in annotators’ cultural and social backgrounds, there is no unified standard for harmful meme definition and classification. 
3) \textbf{Lack of interpretability.} Most datasets provide only binary or multi-class labels, offering little insight into the reasoning behind model decisions. Although some works introduce limited semantic or contextual annotations, they remain insufficient to support interpretability research~\cite{ref52,ref12}.

\begin{figure}[t]
  \centering
  \includegraphics[width=\linewidth]{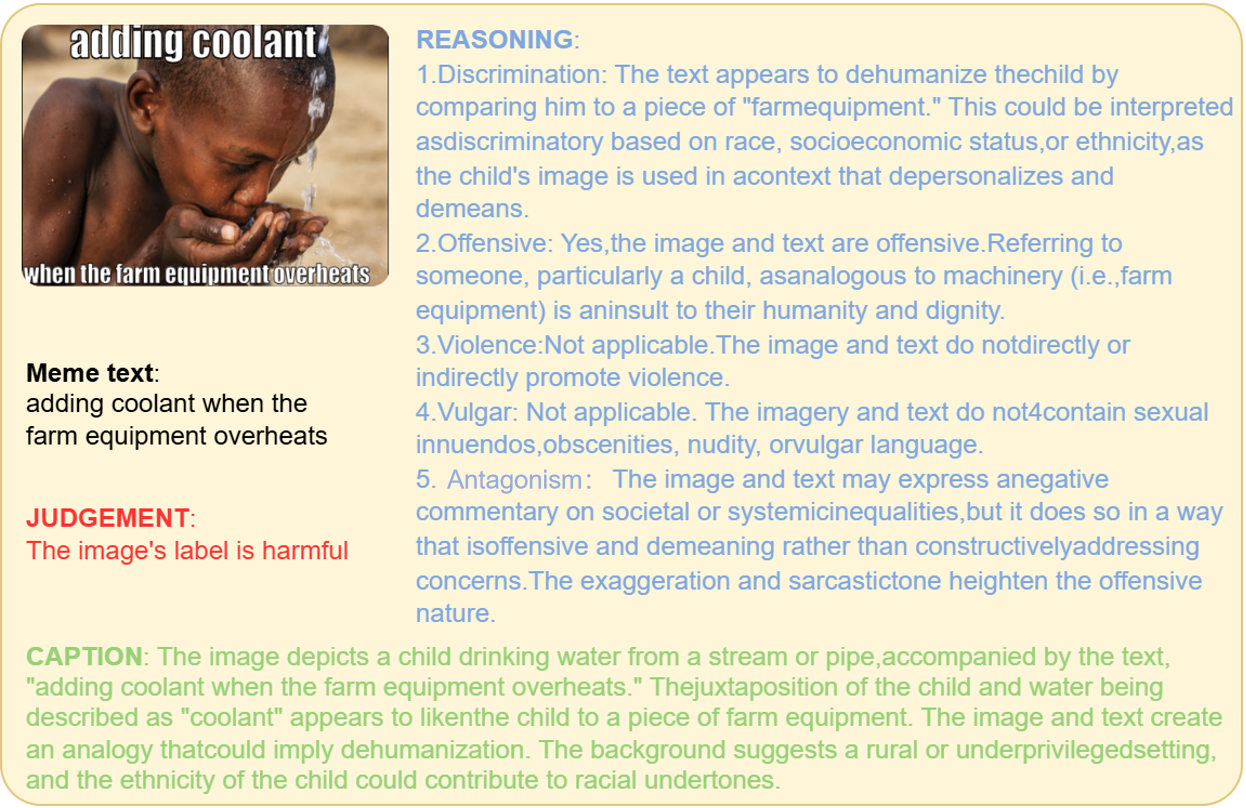}
  \caption{An annotated example from MemeMind. \textbf{CAPTION} interprets the meme, \textbf{REASONING} conducts in-depth analysis, \textbf{JUDGEMENT} is overall harmfulness judgement.}
  \label{example}

\end{figure}

To overcome the limitations of existing datasets, we introduce MemeMind, a new dataset for harmful meme understanding that provides three key advantages:
(1) \textbf{Large scale and diversity.} MemeMind comprises over 40,000 meme samples, approximately three times the size of leading benchmarks such as MAMI~\cite{ref5}. By spanning five harmful categories, MemeMind ensures broad semantic coverage and a relatively balanced binary category distribution.
(2) \textbf{Rigorous and holistic criteria.} We established a systematic classification framework that integrates the internet context with authoritative international standards, such as those from the United Nations Educational, Scientific and Cultural Organization (UNESCO)~\cite{ref62}, the Cyberspace Administration of China (CAC)~\cite{ref64}, and the Organisation for Economic Co-operation and Development (OECD)~\cite{ref63}. This framework standardizes the filtering, annotation, and review workflows, thereby guaranteeing the quality and consistency of the dataset.  
(3) \textbf{Deep semantic annotations.} MemeMind incorporates Chain-of-Thought (CoT)~\cite{Wei2022ChainOT} annotations that simulate human cognitive reasoning, providing robust supervision signals for unraveling the implicit semantics embedded in memes. An example of the annotation is illustrated in Figure~\ref{example}. 

To fully leverage the rich insights in MemeMind, we propose MemeGuard, a multi-stage reasoning enhanced multimodal detection framework. While existing approaches for meme analysis, including traditional black-box detection relying on direct label mapping~\cite{ref3,ref76} or prompt-based LLM methods~\cite{ref8,ref14}, remain constrained by either shallow semantic bridging or surface-level interactions without systematic reasoning, MemeGuard fundamentally breaks through these limitations by introducing multi-stage reasoning enhancement, which has two prominent highlights: (1) \textbf{Semantic reasoning.} Leveraging fine-grained Chain-of-Thought (CoT) annotations, MemeGuard facilitates high-quality reasoning. This CoT-guided mechanism empowers the model to precisely capture multimodal cues and decipher implicit semantics, thereby yielding judgments that are both accurate and interpretable. (2) \textbf{Outstanding performance.}
Leveraging the rich data diversity and high-quality expert annotations provided by MemeMind, MemeGuard achieves superior performance in the harmful meme detection task. Extensive empirical evaluations on the MemeMind benchmark, complemented by cross-domain generalization experiments, consistently demonstrate that MemeGuard significantly outperforms existing multimodal methods.

\begin{figure}[t]
  \centering
  \includegraphics[width=\columnwidth]{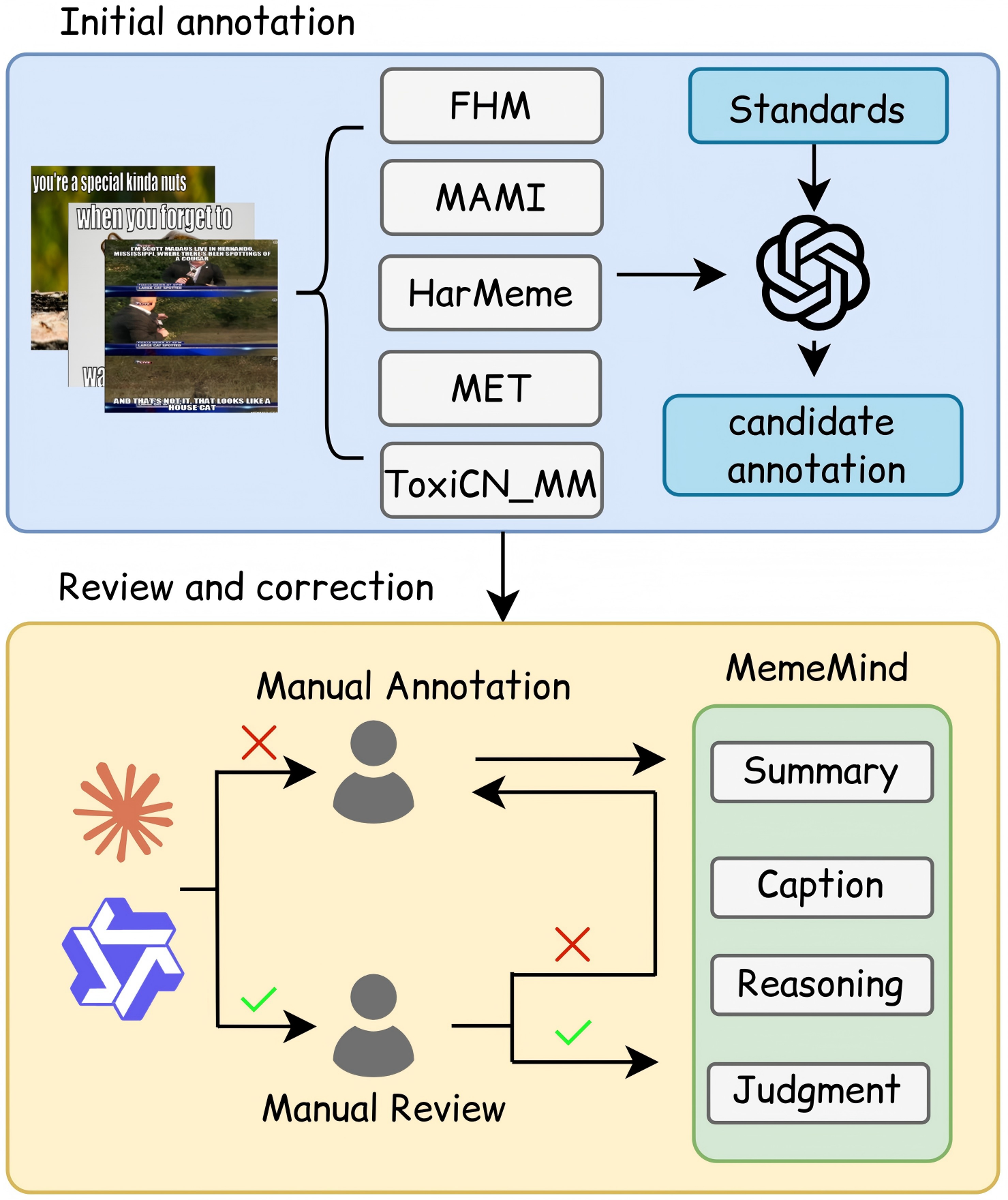}
  \caption{Dataset Construction. We defined scientific standards for harmful meme identification, applied Chain-of-Thought (CoT) annotations simulating human reasoning, implemented a human-centric, model-assisted dual quality control mechanism to ensure dataset quality.}
  \label{Dataset Construction Process}

\end{figure}

The main contributions can be summarized as follows:

\textbullet~We present a large-scale harmful meme dataset, MemeMind, featuring a diverse collection of samples, scientifically rigorous detection criteria, and explainable annotations at the Chain-of-Thought level. 

\textbullet~We propose a reasoning-enhanced harmful meme detection framework, MemeGuard, based on visual language models, which improves the performance and interpretability of harmful meme detection.

\textbullet~We conduct extensive experiments on MemeMind and other datasets, demonstrating that MemeGuard consistently outperforms existing approaches and provides a reliable benchmark as well as a reusable framework for future research.

\section{Related Work}
\subsection{Harmful Meme Datasets.}
Harmful meme detection has emerged as an important research topic for maintaining online content safety and mitigating the societal risks associated with multimodal harmful information. The development of this field has been largely driven by the release of several benchmark datasets. The Facebook Hateful Memes (FHM) dataset~\cite{ref2} is one of the earliest and most influential benchmarks, designed to evaluate whether models can jointly understand visual and textual information rather than relying on unimodal shortcuts. Subsequently, HarMeme~\cite{pramanick-etal-2021-detecting} extended the research scope toward real-world harmful memes, while datasets such as MET~\cite{ref6} introduced broader semantic categories and more diverse multimodal content. These datasets have promoted progress in harmful meme understanding by enriching linguistic expressions, visual themes, and application scenarios. Nevertheless, existing datasets still present several limitations. Many of them are relatively small in scale, focus on a specific type of harmfulness, or adopt task-dependent annotation criteria, making direct comparison across datasets difficult. In addition, most datasets provide only categorical labels or brief textual explanations, without explicitly describing how visual entities, textual expressions, rhetorical devices, and contextual knowledge jointly contribute to harmful meanings. Such limitations restrict the development of models capable of fine-grained reasoning and make it difficult to systematically evaluate the interpretability of harmful meme detection~\cite{ref52,ref12}.

\subsection{Harmful Meme Detection Methods.}
Early harmful meme detection methods primarily relied on the fusion of low-level visual and textual features. Typical approaches extracted textual information using optical character recognition and captured visual semantics through object detection or convolutional neural networks, followed by feature concatenation or attention-based multimodal fusion~\cite{ref3,ref76,ref21,ref22}. Although these methods established effective multimodal baselines, they were often limited in modeling high-level semantic interactions between image and text. With the rapid development of pretrained vision-language models, recent studies have increasingly focused on semantic understanding and knowledge-enhanced reasoning. Representative methods employ Visual Question Answering to decompose meme understanding into a sequence of semantic queries~\cite{ref14,ref15}, use image captioning to generate explicit descriptions of visual content~\cite{ref4,ref7,ref20}, adopt prompting strategies to elicit reasoning capabilities from large multimodal models~\cite{ref8}, or introduce retrieval-enhanced mechanisms to incorporate external knowledge and relevant examples~\cite{ref9}. More recently, ExPO-HM adopts an explain-then-detect paradigm that combines supervised fine-tuning with policy optimization to improve harmful meme detection performance~\cite{ref77}. These advances have significantly improved cross-modal understanding in harmful meme detection. However, harmful meanings are often conveyed through implicit cues such as metaphor, sarcasm, and cultural references. Despite recent progress in reasoning-oriented methods, existing approaches still provide limited supervision for explicitly linking multimodal evidence, harmfulness criteria, and final predictions.


\begin{figure*}[t]
  \centering
  \includegraphics[width=\textwidth]{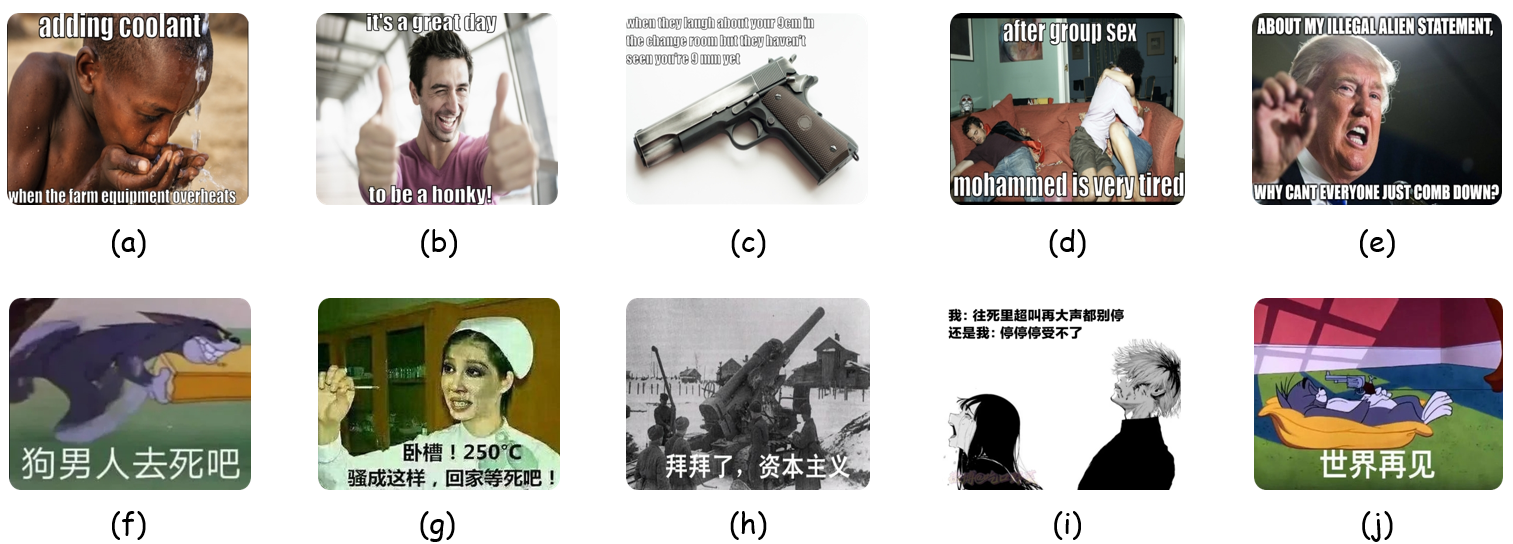}
  \caption{Examples of harmful memes. Images (a) to (e) are English memes, and (f) to (j) are Chinese memes, each corresponding to a specific type of harmful content: (a, f) Discrimination, (b, g) Offensive, (c, h) Violence, (d, i) Vulgar, (e, j) Antagonism.}
  \label{categories}
\end{figure*}

\section{Dataset Construction}

To address the limitations of existing harmful meme datasets in terms of scale, classification standards, and annotation quality, we construct MemeMind based on established content governance principles, leveraging public data sources and a rigorous annotation pipeline. Specifically, we first establish a comprehensive and well-defined harmfulness taxonomy, formulated with reference to international safety and content management standards. Under this framework, we systematically consolidate samples from multiple publicly available harmful meme datasets.
Guided by the harmfulness taxonomy and classification criterion, we further introduce Chain-of-Thought (CoT) annotations to simulate the structured reasoning process of humans in harmfulness assessment, enabling fine-grained classification and reasoning-based labeling. To ensure the rigor and reliability of the annotation work, we implemented a human-centric, model-assisted efficient annotation pipeline. In this framework, large vision-language models are responsible for generating initial candidate reasoning annotations. These annotations are then subjected to rigorous tri-model cross-verification and sampling checks; all samples undergo exhaustive manual review and correction by expert annotators to guarantee that the final labels are grounded in human logic and ethical standards. The pipeline of the dataset construction is illustrated in Figure~\ref{Dataset Construction Process}.

\begin{table*}[t]
\centering
\caption{Illustrations of memes misclassified by the model and manual annotations.}
\label{misclassified_meme}
\renewcommand{\arraystretch}{1.25}
\setlength{\tabcolsep}{5pt}
\small

\resizebox{\linewidth}{!}{%
\begin{tabular}{
  @{}>{\centering\arraybackslash}m{2.8cm}
  >{\centering\arraybackslash}m{3.6cm}
  >{\centering\arraybackslash}m{3.6cm}
  >{\centering\arraybackslash}m{3.6cm}
  >{\centering\arraybackslash}m{3.6cm}@{}
}
\toprule
\textbf{Category} & \textbf{(a)} & \textbf{(b)} & \textbf{(c)} & \textbf{(d)} \\
\midrule

\textbf{Meme Samples} &
\includegraphics[width=\linewidth]{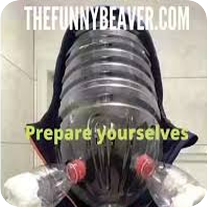} &
\includegraphics[width=\linewidth]{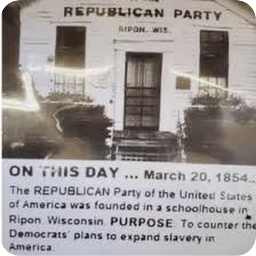} &
\includegraphics[width=\linewidth]{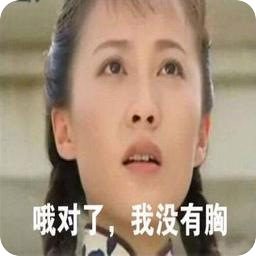} &
\includegraphics[width=\linewidth]{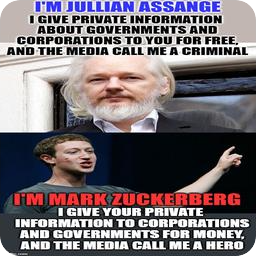} \\
\midrule

\textbf{Model annotation} &
Harmful & Harmful & Nonharmful & Nonharmful \\
\midrule

\textbf{Manual Annotations} &
{\raggedright \textbf{Nonharmful}: uses absurd, self-deprecating humor to mock DIY "protection" with plastic bottles — pure pandemic-era internet fun.\par} &
{\raggedright \textbf{Nonharmful}: factually recalls the GOP's 1854 founding to oppose slavery expansion; neutral and educational, not provocative.\par} &
{\raggedright \textbf{Harmful}: uses the absence of breasts as a punchline, reinforcing body-shaming and gender stereotypes.\par} &
{\raggedright \textbf{Harmful}: contrasts Assange and Zuckerberg to accuse society of hypocrisy over data disclosure, thus politically charged and antagonistic in tone.\par} \\
\bottomrule
\end{tabular}%
}
\end{table*}

\subsection{Definition}
To establish a comprehensive and credible classification standard for harmful content, we integrate multi-level guidelines and systematically design the framework.
First, we referenced the content governance guidelines and international standards issued by CAC~\cite{ref64}, OECD~\cite{ref63} and UNESCO~\cite{ref62} to formulate the initial taxonomy.
Second, we thoroughly analyzed and incorporated the content moderation policies of major social media platforms such as Facebook, Twitter, Xiaohongshu and so on, and refined the framework to ensure both practical applicability and cross-platform consistency.
Finally, after the preliminary framework was developed, we invited experts in content safety to conduct professional review and optimization, further enhancing the classification criteria.
Through the integration of multi-source information and expert participation, we ultimately established a unified, rigorous, and internationally aligned classification system, providing a solid foundation for subsequent data annotation.

Building upon this standardized classification policy, we categorize harmful content into five classes: 
\begin{enumerate}
    \item \textbf{Discrimination:}
    Content that systematically denigrates, excludes, or promotes hostility
    toward individuals or groups based on protected or socially salient attributes
    (e.g., race, ethnicity, gender, religion, disability, LGBTQ+ status, or
    physical appearance). Discrimination typically manifests as derogatory
    stereotypes, dehumanizing language, or calls for unequal treatment.
    The presence of identity-based bias as the primary target or theme is
    the defining criterion.

    \item \textbf{Offensive:}
    Content that insults, mocks, or attacks individuals or groups, showing
    disrespect in tone or attitude but not necessarily involving identity-based
    discrimination. References to sensitive events (e.g., disasters, pandemics,
    or historical tragedies) are considered offensive only when expressed in
    a mocking or disrespectful manner, not for neutral or educational purposes.

    \item \textbf{Violence:}
    Content that explicitly or implicitly encourages, glorifies, or depicts
    physical harm, violent acts, hate or threats thereof. This includes incitement
    to violence, instructions for causing harm, and depictions of violence that
    are not presented in a critical or condemnatory context. Content that merely
    mentions violent events without endorsement or glorification is not
    classified as violence.

    \item \textbf{Vulgar:}
    Content with sexual implications, obscenity, nudity, or crude language,
    including indirect expressions like puns, homophones, or symbolic use of
    objects and gestures to imply sexual acts or organs. Even without explicit
    language, content that clearly evokes sexual associations is considered
    vulgar.

    \item \textbf{Antagonism:}
    Refers to content that utilizes aggressive irony, malicious distortion,
    or extreme exaggeration to deliberately provoke social conflict or
    destabilize public order. Unlike legitimate dissent, this category focuses
    on destructive hostility that transcends factual criticism, aiming to
    incite systemic resentment or polarize communities through inflammatory
    multimodal rhetoric.
\end{enumerate}

\begin{table}[t] 
\centering
\caption{Demographic profile of the five expert annotators.}
\label{tab:annotators_gender_edu}
\small
\renewcommand{\arraystretch}{1.2}
\newcolumntype{S}{>{\hsize=0.7\hsize\centering\arraybackslash}X} 
\newcolumntype{L}{>{\hsize=1.6\hsize\centering\arraybackslash}X} 
\newcolumntype{M}{>{\hsize=0.7\hsize\centering\arraybackslash}X} 

\begin{tabularx}{\columnwidth}{S M L}
\toprule
\textbf{Annotator ID} & \textbf{Gender} & \textbf{Educational Background} \\
\midrule
Annotator A & Female & Master's in Linguistics \\
Annotator B & Male   & Ph.D. in Computer Science \\
Annotator C & Female & Master's in Psychology \\
Annotator D & Male   & Master's in Sociology \\
Annotator E & Female & Bachelor's in Media Studies \\
\bottomrule
\end{tabularx}

\end{table}

\subsection{Data Collection}

We construct MemeMind by consolidating samples from five publicly available harmful meme
datasets, including FHM~\cite{ref2}, HarMeme~\cite{pramanick-etal-2021-detecting}, MAMI~\cite{ref5}, MET~\cite{ref6}, and ToxiCN-MM~\cite{ref4}. After consolidating meme samples from multiple sources, we applied perceptual hashing to identify and remove duplicate instances both within and across source datasets before the train and test split, thereby preventing duplicate samples from being distributed across the two subsets. After deduplication, MemeMind comprises 43,223 meme samples. Rather than directly inheriting the original task-specific labels, all retained samples are re-annotated under our unified harmfulness taxonomy and enriched with structured Chain-of-Thought reasoning annotations. 

The dataset covers a diverse range of harmful content and socially sensitive topics, including offensive content, politics-related content, pandemic-related content, discriminatory content, and pornography. By maintaining a relatively balanced distribution of positive and negative samples, MemeMind achieves broad content coverage and high data quality while preserving a substantial dataset scale. Representative examples are illustrated in Figure~\ref{categories}.

\textbf{Ethical considerations.}
MemeMind is constructed exclusively from existing publicly available datasets. During data collection and integration, we follow the usage policies and redistribution requirements of the corresponding platforms and original data providers. The collected samples are used solely for academic research on harmful meme detection and multimodal content safety. Given that the dataset may contain harmful or otherwise sensitive content, such materials are included only when necessary for research and analysis and do not reflect the views of the authors.

\begin{figure}[t]
  \centering
  \includegraphics[width=\columnwidth]{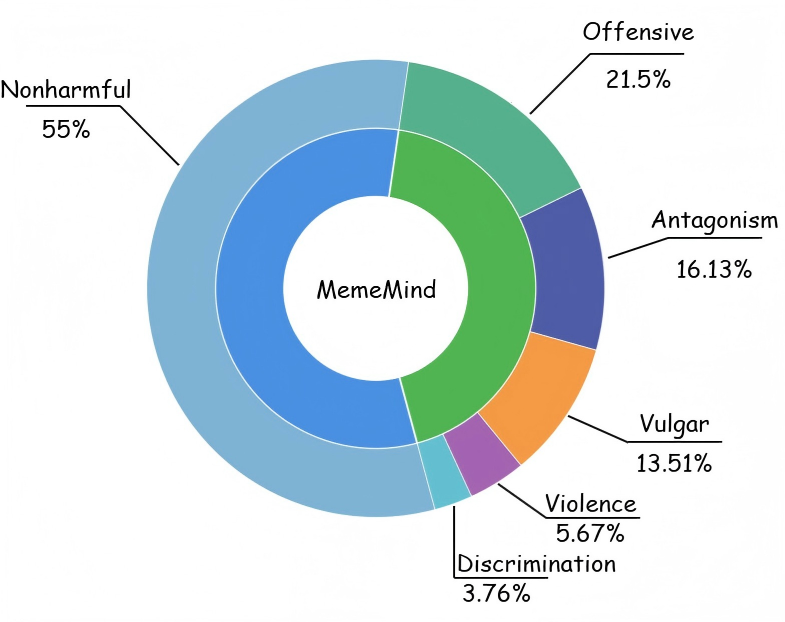}
  \caption{Statistics of dataset distribution. The distribution shows that the ratio of harmful to nonharmful samples is approximately 0.45:0.55. Harmful memes may belong to multiple categories simultaneously and are categorized into the following five non-exclusive subtypes:
\textbf{Discrimination}, \textbf{Offensive}, \textbf{Violence}, \textbf{Vulgar}, and \textbf{Antagonism}.}
  \label{Dataset distrbuted}

\end{figure}

\begin{figure*}[t]
  \centering
  \includegraphics[width=\textwidth]{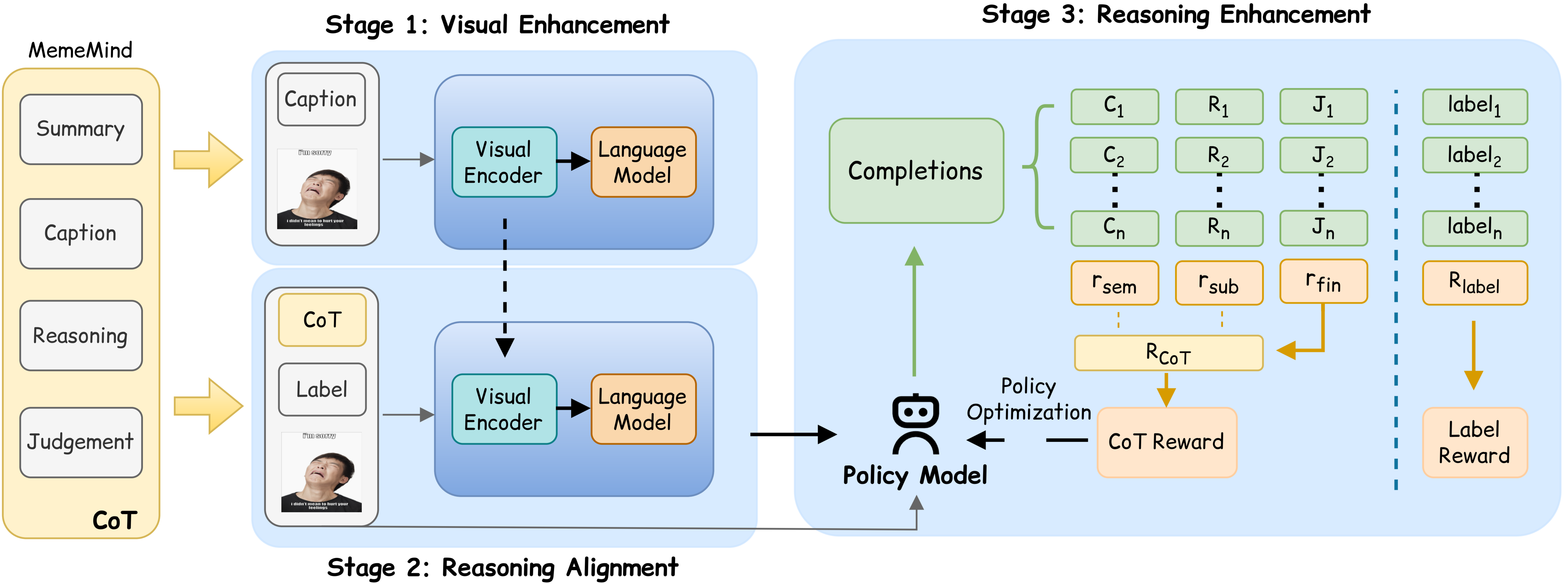}
  \caption{Illustration of \textbf{MemeGuard}.
The framework adopts a three-stage methodology that integrates Chain-of-Thought (CoT) annotations with binary labels. By leveraging Supervised Fine-Tuning and Group Relative Policy Optimization, it systematically establishes and reinforces the model’s reasoning capabilities for harmful meme detection.}
  \label{method}
\end{figure*}

\subsection{Data Annotation}
During the dataset construction process, to ensure rigor, efficiency, and consistency, we established a systematic workflow that integrates model-assisted annotation with manual annotation and review by a professionally trained team. 

\textbf{Initial annotation.} Harmful meme recognition requires not only identifying explicit visual and
textual content, but also interpreting the implicit semantic relationships
between the two modalities and evaluating them against predefined harmfulness
criteria. Following this cognitive process, we formulate harmful meme
annotation as four consecutive stages: (1) specifying the criteria for harmful
content; (2) interpreting the visual and textual information conveyed by the
meme; (3) explicitly comparing the interpreted content against each
harmfulness criterion; and (4) producing the final harmfulness judgment based
on the preceding analysis.

Specifically, each annotation is organized into four structured components:
\textit{SUMMARY}, \textit{CAPTION}, \textit{REASONING}, and
\textit{JUDGMENT}. These components correspond to harmfulness criterion
specification, multimodal semantic interpretation, criterion-wise reasoning,
and final decision making, respectively. By preserving the intermediate
semantic evidence and reasoning basis from content interpretation to label
assignment, this structured format improves the consistency of the annotation
process and provides fine-grained supervision for interpretable harmful meme
recognition. The four components are defined as follows:
\begin{enumerate}
    \item \textit{SUMMARY.}
This component summarizes and specifies the harmfulness dimensions to be examined. Each meme
is evaluated with respect to five predefined categories:
\textit{Discrimination}, \textit{Offensive}, \textit{Violence},
\textit{Vulgar}, and \textit{Antagonism}. For each category, the annotator
determines whether it applies to the current meme. If the content does not
satisfy the corresponding criterion, the category is explicitly marked as
\textit{Not applicable}. By defining a unified set of categories and decision
boundaries, this component establishes a consistent basis for subsequent
semantic interpretation and harmfulness reasoning.
   \item \textit{CAPTION.}
This component provides a comprehensive interpretation of the semantics
conveyed by the meme, rather than merely describing its visible content. The
annotation identifies salient visual entities and textual expressions,
interprets the semantic mappings between the two modalities, and supplements
the contextual knowledge required for understanding the meme. Such information
may include cultural or historical background, character identities,
homophones and wordplay, symbolic expressions, implicit references, and other
context-dependent cues.
   \item \textit{REASONING.}
Based on the multimodal semantic information extracted in the
\textit{CAPTION} component and relevant background knowledge, this component
performs a criterion-wise analysis of the five harmfulness categories. For
each category, the annotation explicitly states whether it is applicable and
provides the corresponding justification. The reasoning process identifies
the supporting visual or textual evidence, explains the semantic relationship
between the modalities, and clarifies how the evidence satisfies or fails to
satisfy the corresponding harmfulness criterion.

  \item \textit{JUDGMENT.}
This component integrates the preceding multimodal interpretation and
category-level reasoning to produce the final binary label, classifying the
meme as either harmful or non-harmful.
\end{enumerate}


\begin{table}[t]
\centering
\caption{Label revision rates under the unified MemeMind taxonomy.}
\label{tab:label_revision}
\small
\renewcommand{\arraystretch}{1.10}
\setlength{\tabcolsep}{4pt}

\begin{tabular*}{\columnwidth}
  {@{\hspace{10pt}}c
   @{\extracolsep{\fill}}c
   @{\hspace{10pt}}}
\toprule
\textbf{Source Dataset}
& \textbf{Label Revision Rate (\%)} \\
\midrule
FHM       & 36.23 \\
MAMI      & 25.30 \\
HarMeme   & 36.54 \\
MET  & 7.47  \\
ToxiCN-MM & 2.26  \\
\bottomrule
\end{tabular*}
\end{table}



Based on the instruction of annotations, to operationalize the proposed annotation paradigm for large-scale dataset
construction, we systematically evaluated multiple candidate
vision-language models and selected GPT-4o~\cite{openai2024gpt4o} as the initial model-assisted
annotation tool. Following the four-stage procedure described above, GPT-4o
generates candidate Chain-of-Thought annotations that include harmfulness
criteria, meme-level semantic interpretation, category-wise reasoning, and
the final judgment. These structured candidates subsequently serve as the
basis for further validation, review, and correction.

\textbf{Review and correction.}
To further ensure the quality of the Chain-of-Thought annotation, we implemented a human-centric, model-assisted dual quality control mechanism. Based on the initial candidate annotations generated by GPT-4o, we introduced Qwen3-VL-Plus~\cite{Bai2025Qwen3VLTR} and Claude 3.5 Sonnet~\cite{anthropic2024claude35sonnet} to perform cross-model consistency verification based on our predefined criteria to mitigate potential single-model biases. 
The cross-model consistency rate reached 88.6\%. To eliminate residual machine bias, human experts conducted a second-pass rigorous audit on this subset, identifying and manually correcting a 16\% error rate to anchor ground-truth reliability. 

For the remaining 11.4\% of samples, we then employed exhaustive manual annotation from scratch by experts, following the same annotation format and standards as the previous subset.
Table~\ref{misclassified_meme} illustrates representative examples of these misclassified memes by models which are corrected by experts. This iterative pipeline resulted in the final MemeMind dataset, featuring high-fidelity Chain-of-Thought reasoning and binary labels for 43,223 samples. Meanwhile, as shown in Table~\ref{tab:label_revision}, re-annotation under the unified MemeMind taxonomy results in label revision rates
ranging from 2.26\% to 36.54\% across the five source datasets.
These differences primarily arise from the heterogeneous
task definitions and labeling criteria adopted by the different datasets.
The detailed distribution of subcategories is visualized in Figure~\ref{Dataset distrbuted}.

\textbf{Dataset quality validation.} To validate the dataset's quality against rigorous statistical standards, we conducted a blind consistency evaluation on a random subset of 1,000 samples involving five trained expert annotators. We employed Fleiss' Kappa~\cite{ref73} to quantify inter-annotator agreement. The overall Fleiss' Kappa was 0.81, with a 95\% confidence interval of [0.78, 0.83], which is conventionally interpreted as ``almost perfect agreement''~\cite{ref74}. To further examine annotation consistency across different harmfulness categories, we additionally sampled 100 instances from each subcategory and obtained Kappa scores of 0.80, 0.82, 0.82, 0.81, and 0.80 for Discrimination, Offensive, Violence, Vulgar, and Antagonism, respectively. These results indicate consistently high inter-annotator agreement across categories and further support the reliability of the MemeMind annotation protocol. Table~\ref{tab:annotators_gender_edu} provides the demographic and professional profiles of the expert annotators.

\section{Methodology}

To address the critical challenge of accurately identifying implicit meme semantics, we propose MemeGuard, a reasoning-oriented framework for harmful meme detection (illustrated in Figure~\ref{method}). The framework achieves high-performance detection through three progressive stages: (1) the \textbf{Visual Enhancement} stage establishes a solid foundation for multimodal understanding; (2) the \textbf{Reasoning Alignment} stage standardizes the model's Chain-of-Thought (CoT) logic; (3) the \textbf{Reasoning Enhancement} stage further refines reasoning quality, thereby enabling accurate and interpretable harmful meme detection.

\subsection{Stage 1: Visual Enhancement}

To bridge the semantic gap between low-level visual content and high-level metaphorical meanings, the first stage of MemeGuard focuses on enhancing the visual understanding capability of the multimodal model.

Specifically, we perform supervised fine-tuning using the detailed caption annotations provided in MemeMind. Given a meme image, the model is trained to generate a comprehensive textual description. By jointly optimizing the visual encoder and language model, the captioning objective encourages the visual encoder to learn more semantically informative representations:
\begin{equation}
\mathcal{L}_{\text{cap}}
=-\sum_{t=1}^{T}\log p_{\theta}
\left(c_t \mid c_{<t}, I\right),
\end{equation}
where $I$ denotes the input meme image, $c_t$ is the $t$-th token of the ground-truth caption, $T$ is the length of the ground-truth caption and $\theta$ represents the trainable model parameters. Through caption-based alignment, the visual encoder is encouraged to transform raw visual signals into semantically meaningful representations that are compatible with the language space of the multimodal model.

\subsection{Stage 2: Reasoning Alignment}

Building upon the Visual Enhancement stage, we conduct targeted harmfulness reasoning fine-tuning to refine the model's reasoning paradigms and bolster its detection efficacy. Each MemeMind sample provides two complementary supervision signals: (1) a Chain-of-Thought (CoT) annotation, which guides the model in learning coherent reasoning trajectories; and (2) a binary harmfulness label, which provides direct supervision for harmfulness classification. 

Specifically, while a standard cross-entropy loss $\mathcal{L}_{\text{cls}}$ governs the classification task based on binary labels, an auxiliary reasoning task is provided by the CoT annotations. Given the input meme $I$ and prompt $P$, the reasoning loss $\mathcal{L}_{\text{CoT}}$ for generating the reasoning is defined as:
\begin{equation}
    \mathcal{L}_{\text{CoT}} = -\frac{1}{T} \sum_{t=1}^{T} \log  p_{\theta}(y_t \mid I, P, y_{<t})
\end{equation}
where $y_t$ denotes the $t$-th token of the ground-truth reasoning sequence, and $T$ is length of the sequence output. Minimizing $\mathcal{L}_{\text{cot}}$ calibrates the model to internalize structured reasoning logic, establishing a reasoning foundation that bridges implicit multimodal semantics with robust classification.

We employ multi-task joint fine-tuning to jointly optimize the classification task and CoT reasoning task, and formulate the unified loss function as:
\begin{equation}
\mathcal{L}_{\text{total}} = \mathcal{L}_{\text{cls}} +  \mathcal{L}_{\text{CoT}}
\label{eq:loss}
\end{equation}
By jointly optimizing the vision encoder and language model through this unified  loss function, we not only achieve alignment in the model's reasoning paradigm but also lay a solid foundation for its reasoning capabilities.

\subsection{Stage 3: Reasoning Enhancement}

Building upon the foundational reasoning capabilities acquired in the previous stages, we further bolster the model through a reinforcement learning paradigm. Unlike conventional Supervised Fine-Tuning, which restricts the model to mimetic learning of pre-defined reasoning steps, Reinforcement Learning facilitates the autonomous exploration of superior logic paths. By employing the Group Relative Policy Optimization (GRPO)~\cite{ref75} framework for multi-task joint fine-tuning, we transition from passive replication to proactive optimization, ensuring more robust, deep, and stable reasoning in harmfulness detection.

For each input prompt $P$ and meme $I$, the model $\pi_\theta$ generates a group of responses $\{o_1, o_2, \dots, o_G\}$, where each $o_i = (Y_i, L_i)$ consists of a reasoning sequence $Y_i$ and a predicted label $L_i$. 
Specifically, we incorporate both types of supervision from the Reasoning Alignment stage and engineer a composite reward mechanism for GRPO training, which contains two reward signals: a CoT reward $R_{\text{CoT}}$ derived from the CoT annotation and a label reward $R_{\text{Label}}$ derived from the binary label annotation.

$R_{\text{CoT}}$ consists of three core components: semantic similarity reward $r_{\text{sem}}$, subcategory harmfulness reward $r_{\text{sub}}$ and overall harmfulness reward $r_{\text{fin}}$, which is defined as:
\begin{equation}
R_{\text{CoT}} = r_{\text{sem}} + r_{\text{sub}} + r_{\text{fin}}
\label{eq:reward}
\end{equation}
For $r_{\text{sem}}$, we extract the meme understanding caption from the generated reasoning chain and employ a combination of BLEU~\cite{ref70}, ROUGE~\cite{ref71}, and BERTScore~\cite{ref72} metrics to evaluate the semantic alignment between the generated and reference captions. This mechanism enhances the model’s capability in multimodal content understanding.
For $r_{\text{sub}}$, we parse the predictions for fine-grained harmful subcategories from the reasoning chain and assign a binary reward to each. The final $r_{\text{sub}}$ is calculated as the mean of these individual rewards. This component promotes model interpretability by providing a reliable, granular basis for the final harmfulness determination.
For $r_{\text{fin}}$, we assess whether the final judgment in the reasoning chain is consistent with the ground-truth judgment and assign a binary reward accordingly. This reward anchors the reasoning process to the correct harmfulness decision, providing a reliable optimization signal.

$R_{\text{Label}}$ evaluates the consistency between the predicted label and the ground-truth label, and assigns a binary reward accordingly, thereby providing a reliable reward signal for the model's harmfulness judgment, which is defined as:
\begin{equation}
R_{\text{Label}} =
\begin{cases}
1, & \text{if } \hat{y} = y \\[6pt]
0, & \text{otherwise}
\end{cases}
\end{equation}

To enhance the model's reasoning capability and effectively leverage the composite reward signals from CoT annotations and binary labels, we optimize MemeGuard using GRPO. For each image--prompt pair, the policy samples a group of G candidate responses. The total reward of each response is computed according to $R_{\text{CoT}}$ and $R_{\text{Label}}$, and the rewards within the group are normalized to derive relative advantages. These advantages are then used to update the policy following the standard GRPO objective, encouraging responses with higher-quality reasoning and more accurate harmfulness judgments while maintaining stable policy optimization. Through this group-relative optimization process, MemeGuard progressively improves the consistency and reliability of its reasoning for harmful content detection.

\begin{table*}[t]
  \centering
    \caption{Performance comparison on the MemeMind dataset. $^{\diamond}$ indicates direct evaluation without fine-tuning. \textbf{Bold} indicates the best result, while \underline{underlining} indicates the second-best result. }
  \label{experiment}
  \small 
  \renewcommand{\arraystretch}{1.5} 
  \setlength{\tabcolsep}{10pt}      

  \begin{tabularx}{\textwidth}{
      >{\centering\arraybackslash\bfseries}m{2.8cm}  
      >{\centering\arraybackslash}X               
      >{\centering\arraybackslash}m{1.8cm}          
      >{\centering\arraybackslash}m{1.8cm}
      >{\centering\arraybackslash}m{1.8cm}
      >{\centering\arraybackslash}m{1.8cm}
  }
    \toprule
    \multirow{2}{*}{\textbf{Category}} & \multirow{2}{*}{\textbf{Method}} & \multicolumn{4}{c}{\textbf{MemeMind}} \\
    \cmidrule(lr){3-6}
    & & \textbf{Accuracy} & \textbf{Precision} & \textbf{Recall} & \textbf{F1-score} \\
    \midrule
    \multirow{2}{*}{\textbf{Single-Modality}} 
      & CLIP Text-Only~\cite{ref53} & 74.14 & 75.54 & 56.92 & 64.92 \\
      & CLIP Image-Only~\cite{ref53} & 80.25 & 79.47 & 71.74 & 75.41 \\
    \midrule
    \multirow{11}{*}{\textbf{Multi-Modality}} 
      & CLIP Image+Text~\cite{ref53} & 80.81 & 80.81 & 71.06 & 75.62 \\
      & VisualBERT\_COCO~\cite{ref18} & 63.21 & 63.95 & 63.48 & 63.72 \\
      & Pro-Cap~\cite{ref13} & 75.58 & 78.09 & 78.76 & 78.44 \\
      & ISSUES~\cite{ref21} & 82.54 & 83.09 & 74.80 & 78.72 \\
      & MemeCLIP~\cite{ref76} & 79.18 & 80.48 & 78.98 & 79.72 \\
      & RGCL~\cite{Mei2023ImprovingHM} & 81.65 & 78.38 & 79.21 & 78.79 \\
      & RA-HMD~\cite{mei-etal-2025-robust} & 81.23 & 80.41 & 74.53 & 77.36 \\
      & Qwen2.5-VL-32B$^{\diamond}$~\cite{ref69} & 78.44 & 81.22 & 63.25 & 71.11 \\
      & Qwen3.6-plus$^{\diamond}$~\cite{qwen36plus} & 77.35 & \textbf{92.26} & 50.98 & 65.67 \\
      & GPT-5.5$^{\diamond}$~\cite{openai2026gpt55} & 77.91 & \underline{86.97} & 57.37 & 69.14\\
      & SFT on MemeMind & \underline{83.93} & 80.48 & \underline{82.73} & \underline{81.59} \\
    \midrule
    \textbf{Our Method} & \textbf{MemeGuard} &  \textbf{87.42} & 83.95 & \textbf{87.51} & \textbf{85.69} \\
    \bottomrule
  \end{tabularx}

\end{table*}


  


 \begin{table}[t]
  \centering
    \caption{Ablation of training data on MemeGuard.}
  \label{data_ablation}

  \small
  \renewcommand{\arraystretch}{1.3} 
  
  \newcolumntype{L}{>{\hsize=1.6\hsize\centering\arraybackslash}X} 
  \newcolumntype{S}{>{\hsize=0.85\hsize\centering\arraybackslash}X} 

  \begin{tabularx}{\columnwidth}{L S S S S}
    \toprule
    \textbf{Training Data} & \textbf{Accuracy} & \textbf{Precision} & \textbf{Recall} & \textbf{F1} \\
    \midrule
    Label Only & 85.01 & 82.96 & 82.01 & 82.49 \\
    \textbf{CoT + Label} & \textbf{87.42} & \textbf{83.95} & \textbf{87.51} & \textbf{85.69} \\
    \bottomrule
  \end{tabularx}

\end{table}

\section{Experiments}

\subsection{Experimental Setup}

\noindent\textbf{Baselines.} In addition to systematically evaluating our enhanced harmful content detection framework MemeGuard on MemeMind, we conducted a comprehensive performance comparison with a variety of baseline models.
The experiments are organized into two groups:
Single Modality Group, where the CLIP model is applied independently to the text and image modalities; MultiModality Group, which includes not only our proposed detection framework but also several representative multimodal methods, such as CLIP~\cite{ref53}, Pro-Cap~\cite{ref13}, ISSUES~\cite{ref21}, U-CoT+~\cite{Pan2025ReadAY}, MemeCLIP~\cite{ref76}, and VisualBERT\_COCO~\cite{ref18}, RA-HMD~\cite{mei-etal-2025-robust}, RGCL~\cite{Mei2023ImprovingHM}, along with several Vision Language Models. Additionally, a foundation model fine-tuned exclusively on MemeMind is also included in this group. To further evaluate the cross-domain generalization capability of MemeGuard, we conduct experiments on PrideMM~\cite{ref76} dataset.

\noindent\textbf{Evaluation strategy.} We employ Accuracy, Precision, Recall, F1-score as the primary evaluation metrics to comprehensively assess the performance of MemeGuard and the baseline methods on the harmful meme detection task.

\noindent\textbf{Implementation details.}
Considering the trade-off between computational efficiency and model performance, we employ Qwen2.5-VL-7B-Instruct~\cite{ref69} as the foundation model. MemeMind is divided into training and test sets at a ratio of 7:3 using stratified sampling, and all baselines follow the same data split unless otherwise specified. We additionally construct a supervised fine-tuning baseline based on Qwen2.5-VL-7B-Instruct, which is trained directly on MemeMind to evaluate the effectiveness of MemeMind. In the Visual Enhancement stage, we perform full-parameter fine-tuning on the training-set caption annotations for one epoch at $1\times10^{-4}$. Reasoning Alignment applies LoRA-based SFT for two epochs at $1\times10^{-4}$, followed by one epoch GRPO at $1\times10^{-5}$ for Reasoning Enhancement, where eight responses are sampled per prompt with a temperature of 1.0. For the latter two stages, the LoRA rank and alpha are both 128. All training uses AdamW, a batch size of 8, and NVIDIA H100 GPUs.

\begin{table}[t]
  \centering
  \caption{Ablation of training stages on MemeGuard.}
  \label{stage_ablation}

  \small
  \renewcommand{\arraystretch}{1.2}

  \newcolumntype{L}{>{\hsize=1.8\hsize\centering\arraybackslash}X}
  \newcolumntype{S}{>{\hsize=0.6\hsize\centering\arraybackslash}X}

  \begin{tabularx}{\columnwidth}{L S S}
    \toprule
    \textbf{Training Configuration} & \textbf{Accuracy} & \textbf{F1} \\
    \midrule
    w/o Stage 1 & 86.81 & 84.63 \\
    w/o Stage 3 & 85.68 & 83.65 \\
    \midrule
    \textbf{MemeGuard} & \textbf{87.42} & \textbf{85.69} \\
    \bottomrule
  \end{tabularx}
\end{table}

\subsection{Experimental Results}

Table~\ref{experiment} presents the performance comparison results of various single-modality methods, multi-modality methods, and Vision Language Models (VLMs) on the MemeMind dataset: 

\textbf{Single-modality methods.}
The text-only CLIP baseline yields the lowest overall performance, achieving
an accuracy of 74.14\% and an F1-score of 64.92\%. This result indicates that
the textual content of a meme alone is often insufficient for reliable
harmfulness identification, as meme text is typically concise, highly
context-dependent, and frequently expressed through sarcasm, euphemism, or
implicit references. The image-only variant improves the accuracy to 80.25\%
and the F1-score to 75.41\%, suggesting that visual content provides
substantial cues for recognizing harmful entities, actions, and contextual
settings. Nevertheless, both variants remain inferior to the stronger
multimodal approaches, demonstrating the importance of jointly modeling the
semantic interaction between the visual and textual components of a meme.

\begin{figure}[t]
    \centering
    \includegraphics[width=\columnwidth]{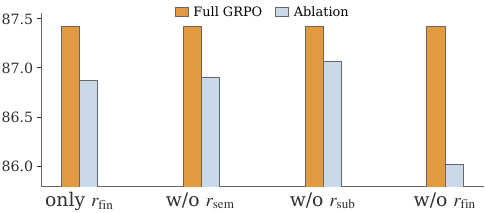}
    \caption{Ablation study of different components in the CoT reward. 
    The performance is measured relative to the full reward design, 
    while the label reward is kept unchanged across all settings.}
    \label{fig:reward_ablation}
\end{figure}

\begin{table}[t]
  \centering
  \caption{Backbone extensibility of MemeGuard.}
  \label{tab:extensibility}

  \small
  \renewcommand{\arraystretch}{1.12}

  \begin{tabular*}{\columnwidth}
    {@{\extracolsep{\fill}} l l c c @{}}
    \toprule
    \textbf{Backbone}
    & \textbf{Setting}
    & \textbf{Accuracy}
    & \textbf{F1} \\
    \midrule

    \multirow{2}{*}{MiniCPM-V-2.6}
    & Zero-shot
    & 73.99
    & 59.94 \\
    & \textbf{MemeGuard}
    & \textbf{87.43}
    & \textbf{85.65} \\
    \midrule

    \multirow{2}{*}{InternVL3-8B}
    & Zero-shot
    & 75.96
    & 75.20 \\
    & \textbf{MemeGuard}
    & \textbf{87.61}
    & \textbf{85.70} \\
    \midrule

    \multirow{2}{*}{Qwen2.5-VL-7B}
    & Zero-shot
    & 71.87
    & 70.57 \\
    & \textbf{MemeGuard}
    & \textbf{87.42}
    & \textbf{85.69} \\
    \bottomrule
  \end{tabular*}

\end{table}

\textbf{Multi-modality methods.}
Most multimodal methods, such as Pro-Cap, ISSUES, MemeCLIP, RGCL, and RA-HMD, outperform the corresponding single-modality
baselines by jointly exploiting visual and textual information.  Among these methods,
MemeCLIP obtains the highest baseline F1-score, whereas ISSUES achieves the
highest baseline accuracy and precision, with values of 82.54\% and 83.09\%.
These results verify the effectiveness of task-specific multimodal
representation learning for harmful meme detection. However, the performance
differences also indicate that existing
fusion-based methods do not consistently balance false-positive and
false-negative errors. Such limitations are particularly relevant for memes
whose harmfulness depends on implicit visual--textual mappings, cultural
knowledge, metaphorical expressions, or context-dependent targets.
Meanwhile, the supervised fine-tuning baseline based on Qwen2.5-VL-7B-Instruct achieves
an accuracy of 83.93\% and an F1-score of 81.59\%, showing that task-specific supervision on MemeMind can effectively adapt a general-purpose VLM to harmful meme detection. Nevertheless, this observation suggests that supervised alignment alone is insufficient to fully capture difficult harmful cases involving implicit or compositional semantics.

\textbf{Directly evaluated VLMs.}
Without task-specific fine-tuning, Qwen2.5-VL-32B~\cite{ref69},
Qwen3.6-plus~\cite{qwen36plus}, and GPT-5.5~\cite{openai2026gpt55}
achieve F1-scores of 71.11\%, 65.67\%, and 69.14\%, respectively.
These results reveal the persistent limitations of current general-purpose
multimodal models in harmful meme detection. They further demonstrate
that relying solely on model scale and general-purpose capabilities is
insufficient to guarantee strong detection performance. Explicit task
alignment and reasoning-oriented optimization remain necessary for
effectively understanding the implicit and complex multimodal semantics
embedded in harmful memes.

In comparison, MemeGuard achieves the best overall performance aggregated over the evaluation metrics,
with an accuracy of 87.42\%, and
an F1-score of 85.69\%. Notably, the
substantial improvement in recall is obtained without sacrificing precision,
resulting in a more balanced identification of harmful and non-harmful memes.
Compared with the supervised fine-tuning baseline built on the same foundation
model and training data, MemeGuard improves accuracy and F1-score by 3.49 and
4.10 percentage points, respectively. This improvement demonstrates the effectiveness of the proposed three-stage training strategy, which progressively enhances multimodal understanding and harmfulness reasoning, leading to stronger overall detection performance.
Overall, these results demonstrate that MemeGuard provides both stronger
classification performance and a more reliable precision--recall trade-off
for harmful meme detection.

\begin{table}[t]
  \centering
  \caption{Cross-domain zero-shot evaluation on PrideMM. \textbf{Bold} indicates the best result, while \underline{underlining} indicates the second-best result.}
  \label{tab:generalization}
  \small
  \renewcommand{\arraystretch}{1.02}
  \setlength{\tabcolsep}{4pt}

  \begin{tabular*}{\columnwidth}
    {@{}l
     @{\extracolsep{\fill}}c
     @{\extracolsep{0pt}\hspace{12pt}}c@{}}
    \toprule
    \textbf{Method}
    & \textbf{Accuracy}
    & \textbf{F1} \\
    \midrule

    \multicolumn{3}{c}{\textit{In-domain Supervised Evaluation}} \\
    \midrule
    CLIP~\cite{ref53}     & 72.40 & 72.30 \\
    MemeCLIP~\cite{ref76} & 76.10 & 75.10 \\
    U-CoT+~\cite{Pan2025ReadAY}   & 71.60 & 71.37 \\
    ISSUES~\cite{ref21}   & 74.68 & 73.64 \\
    RGCL~\cite{Mei2023ImprovingHM}     & 76.30 & 76.50 \\
    RA-HMD~\cite{mei-etal-2025-robust}   & \textbf{78.10} & \textbf{78.40} \\

    \midrule
    \multicolumn{3}{c}{\textit{Cross-domain Zero-shot Evaluation}} \\
    \midrule
    Qwen2.5-VL-7B & 63.12 & 61.12 \\
    SFT on MemeMind & 73.57 & 74.03 \\
    \textbf{MemeGuard} & \underline{76.92} & \underline{78.13} \\
    \bottomrule
  \end{tabular*}
\end{table}

\begin{table}[t]
\centering
\caption{Quantitative Evaluation of Reasoning Quality via Semantic Similarity Metrics.}
\label{tab:text_similarity}
\renewcommand{\arraystretch}{1.3} 
\newcolumntype{Y}{>{\centering\arraybackslash}X} 

\begin{tabularx}{\columnwidth}{@{}lYYY@{}}
\toprule
\textbf{Model} & \textbf{BLEU-4} & \textbf{ROUGE-L} & \textbf{BERTScore} \\ \midrule
Qwen2.5-VL-7B & 12.4 & 18.5 & 0.652 \\
SFT on MemeMind & 32.8 & 42.6 & 0.814 \\
\textbf{MemeGuard} & \textbf{41.2} & \textbf{52.4} & \textbf{0.912} \\ \bottomrule
\end{tabularx}
\end{table}

\subsection{Ablation Study}
In this section, we conduct a series of ablation studies to systematically evaluate the contribution of the Chain-of-Thought (CoT) annotations in MemeMind, the effectiveness of each training stage in MemeGuard, the contribution of individual reward components in Stage 3, and the applicability of the framework across different model backbones. Except for the factor under ablation, all ablation experiments follow the same training configurations described in the preceding sections to ensure fair comparisons.

\begin{table*}[h]
\centering
\caption{Per-category Sub-Acc and overall Decision Alignment (DA) for reasoning sub-categories.}
\label{tab:sub_alignment}
\renewcommand{\arraystretch}{1.2} 
\begin{tabularx}{\linewidth}{c*{5}{>{\centering\arraybackslash}X}c}
\toprule
\multirow{2}{*}{\textbf{Model}} & \textbf{Discrimination} & \textbf{Offensive} & \textbf{Violence} & \textbf{Vulgar} & \textbf{Antagonism} & \multirow{2}{*}{\textbf{DA}} \\
& \textbf{Sub-Acc} & \textbf{Sub-Acc} & \textbf{Sub-Acc} & \textbf{Sub-Acc} & \textbf{Sub-Acc} & \\
\cmidrule(lr){2-6}
Qwen2.5-VL-7B & 52.3 & 60.1 & 58.9 & 61.2 & 59.5 & 62.4 \\
SFT on MemeMind & 78.6 & 84.2 & 81.5 & 83.7 & 82.9 & 84.5 \\
\textbf{MemeGuard} & \textbf{90.5} & \textbf{93.8} & \textbf{91.2} & \textbf{94.1} & \textbf{92.7} & \textbf{95.8} \\
\bottomrule
\end{tabularx}

\end{table*}

\textbf{Ablation on training data:} To evaluate the effectiveness of Chain-of-Thought (CoT) annotations, we maintain the Stage 1 settings and train the model using only binary labels in Stages 2 and 3. This variant is then compared against the full MemeGuard model, which is trained using both CoT annotations and labels.
The results in Table~\ref{data_ablation} clearly demonstrate that when the model is trained solely with binary labels, its F1-score drops notably.
In contrast, incorporating CoT annotations during training leads to a notable performance improvement.
These findings suggest that the supervision provided by Chain-of-Thought (CoT) annotations in MemeMind effectively enhances the model’s performance.

\textbf{Ablation on training configuration:} 
Since Stage 2 aligns the model with the structured CoT format required for reward computation, it is a necessary prerequisite for Stage 3. Removing Stage 2 causes GRPO training to fail to converge; therefore, we keep it fixed and evaluate the contributions of Stages 1 and 3 separately.
To evaluate the contribution of Stage 1, we compare the variant excluding Stage 1 against the full MemeGuard model. The detection results in Table~\ref{stage_ablation} reveal that, compared to the full model, removing Stage 1 leads to a clear decline in the core metrics. This indicates that Stage 1 is pivotal for deepening the semantic understanding of memes.
Furthermore, to assess the efficacy of Stage 3, we compare the variant without Stage 3 against the full MemeGuard. The performance decay upon removing Stage 3 underscores that GRPO is essential for transcending the limitation of supervised learning, thereby regularizing reasoning stability through proactive logic exploration.

\textbf{Ablation on reward:} To investigate the contribution of different reward signals in reasoning optimization of MemeGuard, we conduct an ablation study on the components of $R_{\text{CoT}}$ while keeping all other settings fixed, as illustrated in Fig.~\ref{fig:reward_ablation}. The results show that removing the final judgment reward \(r_{\mathrm{fin}}\) causes the most pronounced deterioration, indicating that outcome-level supervision serves as the primary signal for aligning the reasoning process with the correct harmfulness decision. In contrast, the caption and subcategory rewards provide complementary intermediate supervision by improving multimodal semantic understanding and reasoning quality. Moreover, retaining only the \(r_{\mathrm{fin}}\) still yields inferior performance compared with the complete reward design, suggesting that final-label supervision alone is insufficient for fully optimizing the detection performance.

\textbf{Ablation on model backbones:} To evaluate the extensibility and model-agnostic nature of MemeGuard, we conduct comprehensive experiments across a diverse set of multimodal model backbones. To avoid architectural overlap, we select models from distinct families, including MiniCPM-V-2.6~\cite{Yao2024MiniCPMVAG} and InternVL3-8B~\cite{ref66}. All vanilla backbones are evaluated under a unified zero-shot CoT prompting setting. As illustrated in Table~\ref{tab:extensibility}, integrating MemeGuard consistently improves the performance of all backbones on harmful meme analysis and detection tasks without requiring task-specific architectural modifications. These results validate the robustness of MemeGuard and highlight its broad applicability as a model-agnostic solution for harmful meme analysis and detection.

\subsection{Cross-Domain Experiment} 
To evaluate the cross-domain generalization of MemeGuard and MemeMind, we conduct zero-shot experiments on the PrideMM dataset after training exclusively on MemeMind. As an independently benchmark focusing on LGBTQ+-related harmful memes, PrideMM exhibits substantial topical and semantic differences from MemeMind, providing a suitable out-of-distribution setting for evaluating cross-domain transfer. As shown in Table~\ref{tab:generalization}, the model solely SFT-trained on MemeMind substantially outperforms the original backbone model, and MemeGuard further improves upon this SFT baseline and achieves performance comparable to methods trained directly on PrideMM, thereby demonstrating the promising cross-domain generalization of MemeMind and MemeGuard.

\subsection{Interpretability and Reasoning Evaluation}

To rigorously assess the interpretability of MemeGuard, we perform a quantitative evaluation of the generated Chain-of-Thought (CoT) reasoning. We randomly sampled 400 instances from the test set, comprising 300 harmful and 100 nonharmful samples, using expert-annotated reasoning paths as the ground truth. Our evaluation focuses on two dimensions: (1) \textbf{Semantic Similarity}, which measures the linguistic alignment of caption, and (2) \textbf{Sub-category Reasoning Accuracy}, which evaluates the model's ability to correctly identify specific harmful logic.

\textbf{Evaluation of semantic similarity.} 
We employ a suite of standard Natural Language Processing metrics to quantify the similarity between model-generated reasoning and expert annotations. These include BLEU-4, ROUGE-L for n-gram overlap, as well as BERTScore for deep semantic similarity.

As shown in Table~\ref{tab:text_similarity}, MemeGuard consistently outperforms the base model and the SFT baseline across all evaluation metrics, indicating improved reasoning quality and interpretability. These results suggest that the proposed training framework enables the model to generate reasoning outputs that are better aligned with the underlying harmful semantics of memes.

\begin{table*}[t]
\centering
\caption{Representative false-positive case of MemeGuard.
The model correctly interprets the multimodal content but
incorrectly applies the harmfulness criteria. Only the reasoning
components directly related to the error are presented.}
\label{tab:failure_case}

\footnotesize
\setlength{\tabcolsep}{6pt}
\renewcommand{\arraystretch}{1.08}

\begin{tabularx}{\textwidth}{
  @{}
  P{0.18\textwidth}
  Y
  Y
  @{}
}
\toprule

\rowcolor{gray!12}
\centering\arraybackslash\textbf{Input Meme}
&
\centering\arraybackslash\textbf{MemeGuard Output}
&
\centering\arraybackslash\textbf{Expert Annotation}
\\

\midrule

\multirow[t]{3}{0.18\textwidth}{%
  \begin{minipage}[t]{\linewidth}
  \vspace{40pt}
  \centering

  \includegraphics[
    width=0.88\linewidth
  ]{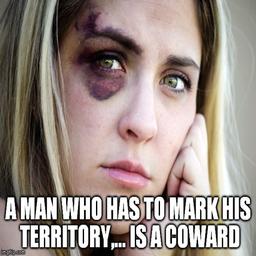}

  \vspace{15pt}

  \begin{minipage}{0.88\linewidth}
  \raggedright

  \textbf{Prediction:} Harmful

  \vspace{5pt}

  \textbf{Ground Truth:} Non-harmful

  \vspace{2pt}

  \end{minipage}
  \end{minipage}
}
&
\textbf{\textsc{Caption}}\par
\vspace{2pt}

The image shows a woman with visible bruises around her eyes,
suggesting physical assault. The accompanying text reads,
``A man who has to mark his territory \ldots{} is a coward.''
The text criticizes the man's alleged abusive behavior.

&
\textbf{\textsc{Caption}}\par
\vspace{2pt}

The image shows a woman with a visible black eye and the
statement, ``A man who has to mark his territory \ldots{} is
a coward.'' The phrase ``mark his territory'' metaphorically
refers to possessive and physically abusive behavior in an
intimate relationship. The meme adopts a condemnatory stance
toward domestic violence.
\\[4pt]

\cmidrule(lr){2-3}

&
\textbf{\textsc{Reasoning}}\par
\vspace{2pt}

\textit{\textbf{Offensive: }}
The meme references domestic violence, which is a sensitive
and serious issue. Its expression may trivialize the experiences
of victims and is therefore considered offensive.

\vspace{4pt}

\textit{\textbf{Violence: }}
The meme depicts the aftermath of physical assault and
indirectly normalizes violence, which may desensitize viewers
to the seriousness of domestic abuse.

\vspace{4pt}

\textit{\textbf{Antagonism: }}
The meme expresses strong negativity and resentment toward
the alleged perpetrator through aggressive criticism.

&
\textbf{\textsc{Reasoning}}\par
\vspace{2pt}

\textit{\textbf{Offensive:}}
The meme criticizes abusive behavior and raises awareness of
domestic violence rather than mocking the victim or
trivializing the incident.

\vspace{4pt}

\textit{\textbf{Violence:}}
Although the image depicts the aftermath of physical violence,
the injury is presented in a critical and condemnatory context.
The meme neither encourages, glorifies, nor normalizes violence.

\vspace{4pt}

\textit{\textbf{Antagonism: }}
The negative expression targets abusive behavior and functions
as constructive social criticism rather than malicious
provocation or an attempt to intensify social hostility.
\\[4pt]

\cmidrule(lr){2-3}

&
\textbf{\textsc{Judgment}}\par
\vspace{2pt}

\textbf{Harmful}

&
\textbf{\textsc{Judgment}}\par
\vspace{2pt}

\textbf{Non-harmful}
\\[3pt]

\bottomrule
\end{tabularx}
\end{table*}

\begin{table*}[t]
\centering
\caption{Representative false-negative case of MemeGuard.
The model incorrectly interprets the ambiguous visual gesture
as an expression of affection and consequently fails to
recognize the sexual double entendre. Only the reasoning
components directly related to the error are presented.}
\label{tab:failure_case_image_1311}

\footnotesize
\setlength{\tabcolsep}{6pt}
\renewcommand{\arraystretch}{1.08}

\begin{tabularx}{\textwidth}{
  @{}
  P{0.18\textwidth}
  Y
  Y
  @{}
}
\toprule

\rowcolor{gray!12}
\centering\arraybackslash\textbf{Input Meme}
&
\centering\arraybackslash\textbf{MemeGuard Output}
&
\centering\arraybackslash\textbf{Expert Annotation}
\\

\midrule

\multirow[t]{3}{0.18\textwidth}{%
  \begin{minipage}[t]{\linewidth}
  \vspace{20pt}
  \centering

  \includegraphics[
    width=0.88\linewidth
  ]{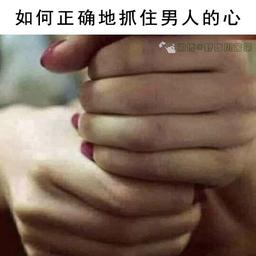}

  \vspace{10pt}

  \begin{minipage}{0.88\linewidth}
  \raggedright

  \textbf{Prediction:} Non-harmful

  \vspace{5pt}

  \textbf{Ground Truth:} Harmful

  \vspace{2pt}

  \end{minipage}
  \end{minipage}
}
&
\textbf{\textsc{Caption}}\par
\vspace{2pt}

The text translates to ``How to correctly grab a man's
heart.'' The image shows a close-up of two hands clasping
around each other, symbolizing affection or emotional
connection. The context appears lighthearted and romantic,
focusing on interpersonal relationships.

&
\textbf{\textsc{Caption}}\par
\vspace{2pt}

The image contains the text ``How to correctly grab a man's
heart'' and depicts a close-up of an ambiguous grasping
gesture. The visual gesture and textual metaphor jointly
create a sexual double entendre. Although the meme maintains
plausible deniability at the literal level, its composition
implicitly evokes a sexual organ or sexual act.
\\[4pt]

\cmidrule(lr){2-3}

&
\textbf{\textsc{Reasoning}}\par
\vspace{2pt}

\textit{\textbf{Vulgar:}}
The content does not contain sexual implications, obscenity,
or crude language. The clasped hands convey a positive and
lighthearted message about forming an emotional bond.
Therefore, this category is not applicable.

&
\textbf{\textsc{Reasoning}}\par
\vspace{2pt}

\textit{\textbf{Vulgar:}}
The metaphor and visual elements strongly imply sexual acts
or organs indirectly. In particular, the grasping gesture
acquires a suggestive meaning when combined with the phrase
``grab a man's heart.'' This cross-modal double entendre
evokes an inappropriate sexual association even though no
explicit sexual language or nudity is present. Therefore,
this category is applicable.
\\[4pt]

\cmidrule(lr){2-3}

&
\textbf{\textsc{Judgment}}\par
\vspace{2pt}

\textbf{Non-harmful}

&
\textbf{\textsc{Judgment}}\par
\vspace{2pt}

\textbf{Harmful}
\\[3pt]

\bottomrule
\end{tabularx}
\end{table*}

\textbf{Sub-category reasoning alignment.} A critical aspect of reasoning interpretability is the correct identification of harmful sub-types. We extract the specific harmfulness sub-categories mentioned within the generated CoT and compare them against the expert labels. We define two key metrics: (1) Sub-Acc: The accuracy of identifying the correct harmful sub-category within the reasoning path. (2) Decision-Alignment (DA): The consistency between the identified sub-category logic and the final harmfulness judgment.


The results in Table~\ref{tab:sub_alignment} demonstrate the superior interpretability of MemeGuard, achieving Sub-Acc above 90\% across all sub-categories and a Decision-Alignment rate of 95.8\%. These results indicate improved consistency between sub-category reasoning and final harmfulness judgments, further supporting the effectiveness of the Reasoning Enhancement stage.

\textbf{Failure case study.} 
We analyze a representative failure case of MemeGuard.
As shown in Table~\ref{tab:failure_case}, MemeGuard correctly
recognizes the physical injury depicted in the image, but still produces a false-positive
prediction. Specifically, MemeGuard incorrectly interprets the
depiction and condemnation of the consequences of violence as
an endorsement of violent behavior, while misclassifying the
criticism of such behavior as offensive or antagonistic
expression. In contrast, the expert annotation classifies the meme
as non-harmful because its primary intent is to raise public
awareness of domestic violence and condemn abusive behavior.
This case indicates that MemeGuard still requires stronger
modeling of communicative stance and more fine-grained decision
boundaries between the endorsement of harmful behavior and
critical expressions directed against such behavior.

Beyond such stance-related false positives, we further analyze a representative false negative case, as shown in Table~\ref{tab:failure_case_image_1311}. Although MemeGuard correctly parses the textual statement “How to correctly grab a man's heart”, it interprets the gesture in the image as ordinary hand‑holding that symbolizes affection and emotional connection. This literal, benign interpretation causes the subsequent reasoning pipeline to overlook the implicit sexual connotation and misclassify the meme as non‑harmful. In contrast, the expert annotation
identifies a cross-modal double entendre: the ambiguous grasping
gesture, when combined with the textual metaphor, indirectly evokes
a sexual organ or sexual act and therefore constitutes vulgar and
harmful content. This case demonstrates that once a benign interpretation is prematurely assigned to the gesture, the error propagates along the reasoning chain and yields a false‑negative result. It also reveals that the model still has room for improvement in recognizing visual metaphors and implicit sexual semantics.

\section{Conclusion}
In this work, we present MemeMind, the first large-scale harmful meme dataset constructed with rigorous and systematic classification standards and enriched with Chain-of-Thought annotations. The dataset is collected and annotated in accordance with international standards and mainstream social media platform guidelines, comprising over 40,000 samples spanning diverse harmful categories, laying a solid foundation for interpretable and robust harmful meme detection. To fully leverage the potential of MemeMind, we propose MemeGuard, a reasoning-enhanced detection framework that employs a three-stage training strategy, improving harmful meme detection performance. Extensive experiments on MemeMind and cross-domain dataset demonstrate that MemeGuard achieves superior performance compared to state-of-the-art baselines. We believe this reasoning-oriented paradigm offers a promising direction for future research in the field.

\bibliographystyle{IEEEtran}
 \bibliography{reference}

 \vspace{-6ex}
\begin{IEEEbiographynophoto}{Hexiang Gu}
received the B.E. degree from Beijing University of Posts and Telecommunications, Beijing, China, in 2025.
He is currently pursuing the Ph.D. degree at the School of Artificial Intelligence, Beijing University of Posts and Telecommunications.
His research interests include multimodal large language model, and multimodal content safety.
\end{IEEEbiographynophoto}

 \vspace{-6ex}
\begin{IEEEbiographynophoto}{Qifan Yu}
received the B.E. degree from Beijing University of Posts and Telecommunications, Beijing, China, in 2025.
He is currently pursuing the M.S. degree at the School of Artificial Intelligence, Beijing University of Posts and Telecommunications.
His research interests include multimodal large language model, and multimodal content safety.
\end{IEEEbiographynophoto}

 \vspace{-6ex}
\begin{IEEEbiographynophoto}{Yuan Liu}
received the B.E. degree from Shandong University, Shandong, China, in 2024.
He is currently pursuing the M.S. degree at the School of Artificial Intelligence, Beijing Normal University.
His research interests include multimodal large language models and multimodal content understanding.
\end{IEEEbiographynophoto}

 \vspace{-6ex}
\begin{IEEEbiographynophoto}{Zikang Li}
received the B.E. degree from Beijing University of Posts and Telecommunications, Beijing, China, in 2024.
He is currently pursuing the Ph.D. degree at Beijing University of Posts and Telecommunications, Beijing, China, as part of a joint doctoral training program with Zhongguancun Academy.
His research interests include reinforcement learning, multi-agent reinforcement learning, and game artificial intelligence, with a focus on robust evaluation, out-of-distribution opponent generalization, and intelligent agents in complex games.
\end{IEEEbiographynophoto}

\vspace{-6ex}
\begin{IEEEbiographynophoto}{Saihui Hou}
received the B.E. and Ph.D. degrees from the University of Science and Technology of China in 2014 and 2019, respectively.
He is currently an Associate Professor at the School of Artificial Intelligence, Beijing Normal University.
His research interests include computer vision and machine learning, with a focus on gait recognition, which aims to identify individuals based on their walking patterns.
\end{IEEEbiographynophoto}

\vspace{-6ex}
\begin{IEEEbiographynophoto}{Jian Zhao}
received the B.E. and Ph.D. degrees in engineering from the University of Science and Technology of China, Hefei, China, in 2018 and 2023, respectively.
He is currently an Assistant Professor and Ph.D. Supervisor with the Zhongguancun Academy, Beijing, China, and with the Zhongguancun Institute of Artificial Intelligence. He has authored/co-authored more than 40 papers in top-tier international conferences and journals, including NeurIPS, ICML, and ICLR.
\end{IEEEbiographynophoto}

\vspace{-6ex}
\begin{IEEEbiographynophoto}{Zhaofeng He}
received the B.S. degree in Electronic Engineering and Information Science from the University of Science and Technology of China, Hefei, China, in 2005, and the Ph.D. degree in Computer Applied Technology from the National Laboratory of Pattern Recognition, Institute of Automation, Chinese Academy of Sciences, Beijing, China, in 2010. He is currently a Professor at the School of Artificial Intelligence, Beijing University of Posts and Telecommunications, Beijing. His research interests include computer vision, biometrics, and intelligent decision-making in games.
\end{IEEEbiographynophoto}

\end{document}